\pdfoutput=1

\documentclass[11pt]{article}

\usepackage[final]{acl}

\usepackage{times}
\usepackage{latexsym}
\usepackage{amsmath}

\usepackage{graphicx}
\usepackage{booktabs}
\usepackage{tabularx}
\usepackage{tcolorbox}
\usepackage{float}

\usepackage[T1]{fontenc}

\usepackage[utf8]{inputenc}

\usepackage{microtype}

\usepackage{inconsolata}

\usepackage{graphicx}

\title{TheoremExplainAgent: Towards Video-based Multimodal Explanations for LLM Theorem Understanding}

\newcommand{\aspace}{\hspace{0.75em}}

\newcommand{\uwaterloo}{$^{\spadesuit}$}
\newcommand{\votee}{$^{\heartsuit}$}
\newcommand{\vectorins}{$^{\dagger}$}

\author{
\uwaterloo \vectorins Max Ku$^*$ \aspace \votee Thomas Chong$^*$ \aspace \uwaterloo Jonathan Leung \aspace \uwaterloo Krish Shah\aspace \votee Alvin Yu \aspace \uwaterloo \vectorins Wenhu Chen
\\
{\small{\texttt{m3ku@uwaterloo.ca, thomas.chong@votee.ai, wenhu.chen@uwaterloo.ca} }} \\ 
\uwaterloo University of Waterloo \aspace  \votee Votee AI \aspace \vectorins Vector Institute  \quad  \\
}

\begin{document}
\twocolumn[{%
    \renewcommand\twocolumn[1][]{#1}%
    \maketitle
    \centering
    \vspace{-10mm}
    \url{https://tiger-ai-lab.github.io/TheoremExplainAgent/}
    \vspace{2mm}
    \begin{center}
        \centering
        \includegraphics[width=0.95\textwidth]{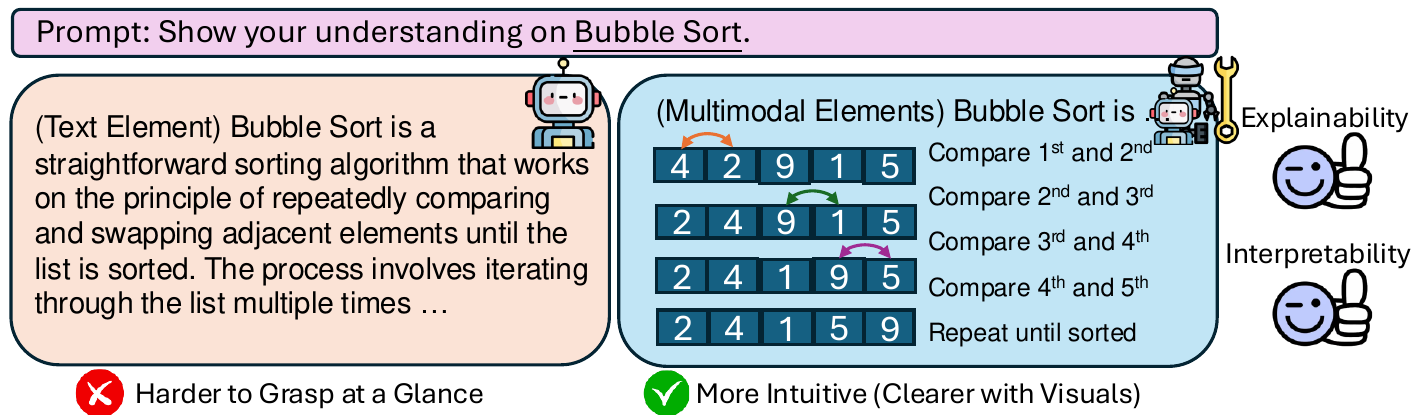}
        \captionof{figure}{We do not have knowledge of a thing until we have grasped its cause~\cite{1901aristotle}. A strong reasoning model should not only generate correct conclusions but also communicate them effectively. Visualization enhances human intuition by making abstract concepts more concrete and revealing hidden relationships. Moreover, visual explanations expose reasoning errors more clearly than text, making it easier to diagnose model mistakes.}
        \label{fig:teaser}
    \end{center}
}]

\begin{abstract}
Understanding domain-specific theorems often requires more than just text-based reasoning; effective communication through structured visual explanations is crucial for deeper comprehension. While large language models (LLMs) demonstrate strong performance in text-based theorem reasoning, their ability to generate coherent and pedagogically meaningful visual explanations remains an open challenge. In this work, we introduce TheoremExplainAgent, an agentic approach for generating long-form theorem explanation videos (over 5 minutes) using Manim animations. To systematically evaluate multimodal theorem explanations, we propose TheoremExplainBench, a benchmark covering 240 theorems across multiple STEM disciplines, along with 5 automated evaluation metrics. Our results reveal that agentic planning is essential for generating detailed long-form videos, and the o3-mini agent achieves a success rate of 93.8\% and an overall score of 0.77. However, our quantitative and qualitative studies show that most of the videos produced exhibit minor issues with visual element layout. Furthermore, multimodal explanations expose deeper reasoning flaws that text-based explanations fail to reveal, highlighting the importance of multimodal explanations.

\end{abstract}

\section{Introduction}

\begin{figure*}[!t]
    \centering
    \includegraphics[width=0.95\linewidth]{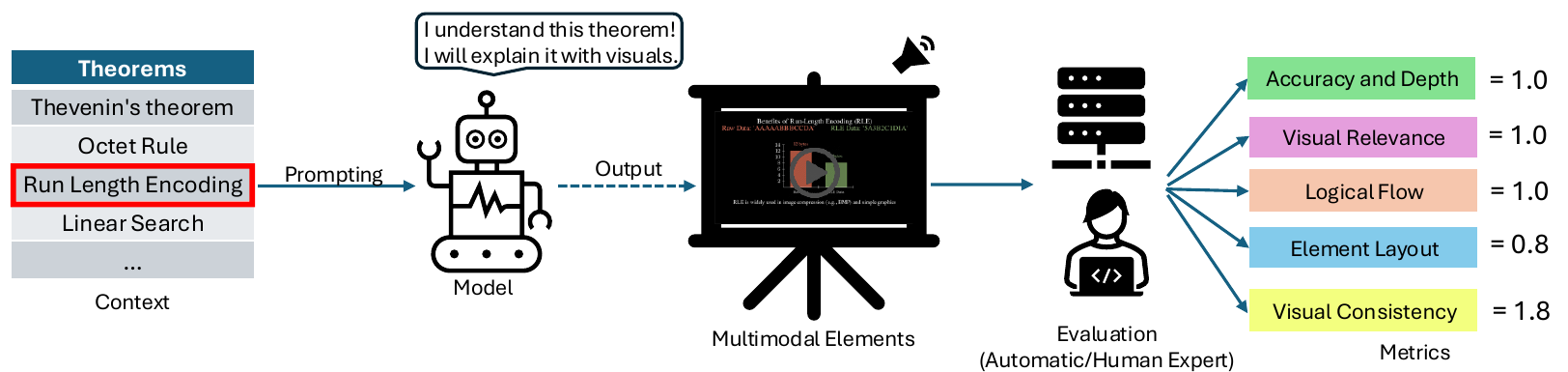}
    \caption{An overview of the multimodal theorem explanation framework.}
    \label{fig:framework_main}
\end{figure*}

A key objective of AI systems is to assist humans in solving complex problems, particularly in domain-specific challenges. To achieve this, AI must go beyond surface-level pattern matching to achieve deeper conceptual understanding to effectively address these problems. Recent research has proposed evaluating AI performance on theorem-driven datasets through multiple-choice question answering~\cite{zhang2024multiplechoicequestionsefficientrobust} and open-ended short question answering~\cite{chen2023theoremqa}. However, these approaches primarily assess textual reasoning and may not fully capture an AI system's ability to grasp theorem concepts at a deeper level. Studies have shown that AI models can be sensitive to superficial cues, such as the order of answer choices in multiple-choice questions~\cite{pezeshkpour2023largelanguagemodelssensitivity, keluskar2024llmsunderstandambiguitytext}. This raises concerns about the robustness of such evaluations in truly measuring comprehension. Moreover, current theorem-focused datasets are predominantly text-based, overlooking how complex concepts are often best understood through structured visualizations.

Theorem reasoning is inherently multimodal, particularly in areas such as geometry, topology, and certain aspects of algebra, where visual representations and spatial reasoning play a crucial role in understanding structures and proving properties. Cognitive science research suggests that multimodal elements improve conceptual understanding, aiding in the comprehension of abstract ideas. Although some studies leverage multimodal input to improve AI reasoning~\cite{zhang2023multicot}, currently there is no standardized evaluation framework to evaluate AI's ability to generate multimodal explanations for complex concepts, which would require models to express knowledge in an interpretable manner. This raises the question: \textbf{Can AI systems effectively generate multimodal theorem explanations?} 

As video is a classic example of multimodal data, we explore the question by introducing TheoremExplainAgent, an agentic AI system designed to generate theorem explanations in the form of explanatory videos. TheoremExplainAgent demonstrates the capability to plan and generate long, coherent videos by mimicking human video production processes. In this system, a planner agent generates story plans and narrations, and a coding agent generates Python animation scripts using Manim~\cite{The_Manim_Community_Developers_Manim_Mathematical_2024} to create long and meaningful videos.
Additionally, to systematically evaluate AI-generated explanations, we develop TheoremExplainBench, a benchmark suite comprising 240 theorems spanning four STEM disciplines. We assess AI-generated explanations based on 5 dimensions related to factual correctness and perceptual quality, using automatic or human-evaluation metrics. An overview of the framework is illustrated in Figure~\ref{fig:framework_main}. 

Our experiments with TheoremExplainAgent yielded both promising results and clear areas for improvement in AI-generated multimodal theorem explanations. On the positive side, a key achievement was the system's ability to generate extended video explanations, reaching durations of up to 10 minutes. This represents a significant advancement over agentless approaches, which we found to be limited to approximately 20-second videos. Furthermore, TheoremExplainAgent demonstrated versatility across different STEM disciplines, successfully creating videos for Mathematics, Physics, Chemistry, and Computer Science. Importantly, we observed that video-based theorem explanations inherently expose deeper reasoning flaws in AI systems that text-based evaluations often miss. Unlike text-based multiple-choice questions, where models can exploit superficial cues, generating visual-theorem explanations necessitates that the AI explicitly encodes structural and procedural knowledge, thus making underlying errors more apparent. In particular, the o3-mini model exhibited robust performance at varying levels of theorem difficulty, indicating a capacity to handle both fundamental and complex concepts. However, despite these successes, limitations persist. While the system could generate textually accurate explanations, the visual quality and pedagogical structure of the videos require further refinement. Generated animations frequently exhibited minor visual layout inaccuracies, such as misaligned text elements, overlapping shapes, and inconsistent object sizes. These visual errors, though often subtle, became more pronounced and potentially distracting. This happens particularly in the medium and hard difficulty levels of our TheoremExplainBench.

\begin{figure*}[!t]
    \centering
    \includegraphics[width=1\linewidth]{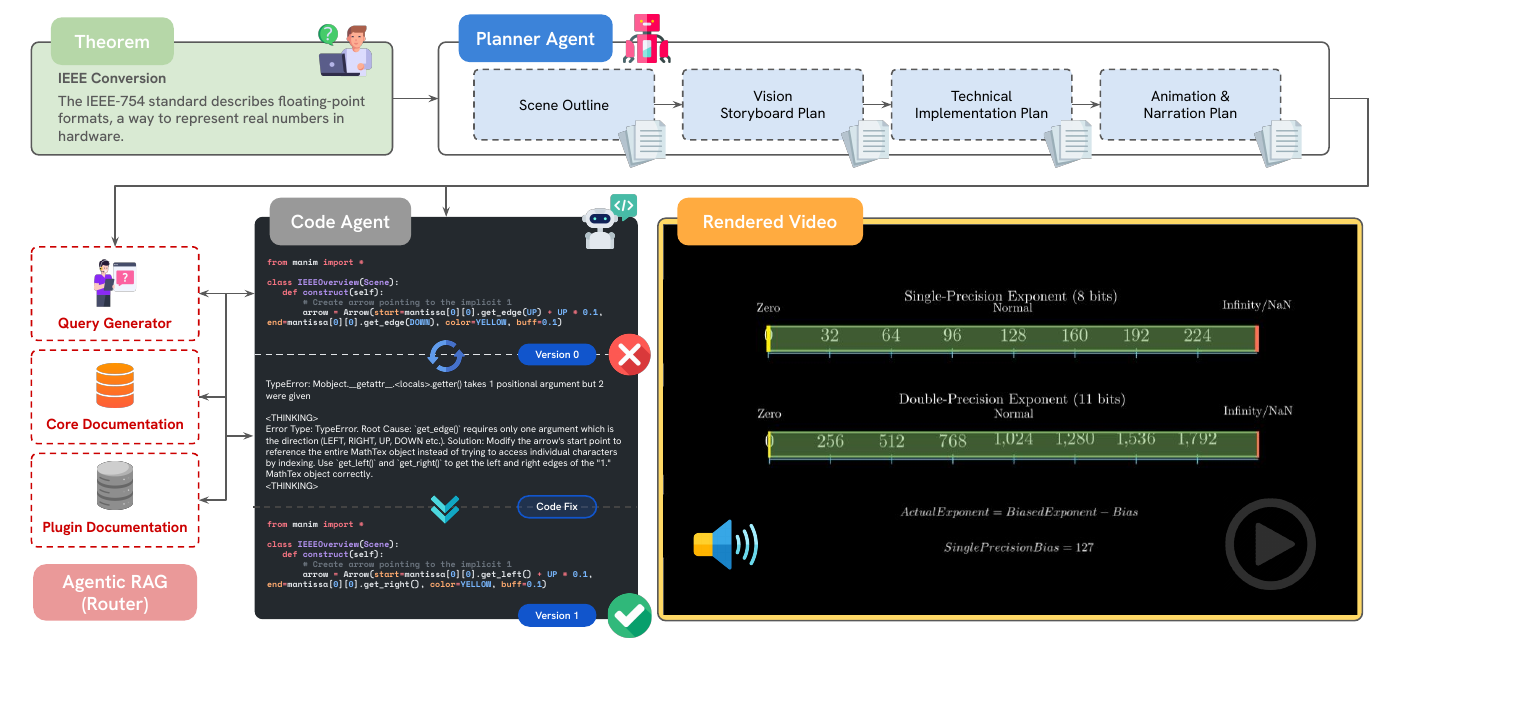}
    \caption{TheoremExplainAgent consists of two LLM agents. Taking a theorem as input, the planner agent create plans for execution. The coding agent then generates Python scripts to produce visuals and audio.}
    \label{fig:method_main}
\end{figure*}

Therefore, the major contributions of this work: 

(1) Task Definition. We introduce the novel problem of AI-generated multimodal theorem explanations and identify the key challenges associated.

(2) TheoremExplainAgent. We develop an agentic approach to generating explanatory videos, as a baseline to assess current AI capabilities.

(3) TheoremExplainBench. We curate a diverse benchmark dataset spanning 4 STEM disciplines and propose 5 automatic evaluation metrics, measuring progress toward solving this problem.

\section{Related Works}

\subsection{LLM and Agents}
The rapid advancements in large language models (LLMs) and large vision-language models (VLMs) have unlocked unprecedented capabilities in understanding multimodal content. Models such as GPT-4~\cite{openai2023gpt4}, Gemini~\cite{geminiteam2024gemini}, Claude-3.5 Sonnet v1~\cite{anthropic2025claude3}, and DeepSeek~\cite{deepseekai2024deepseekv3technicalreport} have demonstrated strong abilities in processing complex textual information and analyzing visual inputs within a unified framework~\cite{zhang2023multicot}. These breakthroughs have enabled transformative applications across various domains, including visual content understanding~\cite{hu2023tifa, ku2023viescore}, code generation~\cite{nijkamp2023codegenopenlargelanguage, jimenez2024swebench, yang2024swebenchmultimodalaisystems}, and reasoning over structured data. To tackle complex tasks, researchers have explored LLM agents: AI systems that leverage LLMs to autonomously reason, plan, and execute tasks by interacting with structured environments or external tools. These agents have been deployed in various goal-oriented applications, such as scientific discovery~\cite{lu2024aiscientist, si2024llmsgeneratenovelresearch, schmidgall2025agentlaboratoryusingllm}, coding solutions~\cite{abramovich2024enigmaenhancedinteractivegenerative}, multimodal visual generation~\cite{he2024kubrickmultimodalagentcollaborations}, and computer environment interaction~\cite{OSWorld}. In this work, we extend the use of LLM agents into the domain of theorem explanation and visualization.

\begin{figure*}[!t]
    \centering
    \includegraphics[width=0.85\linewidth]{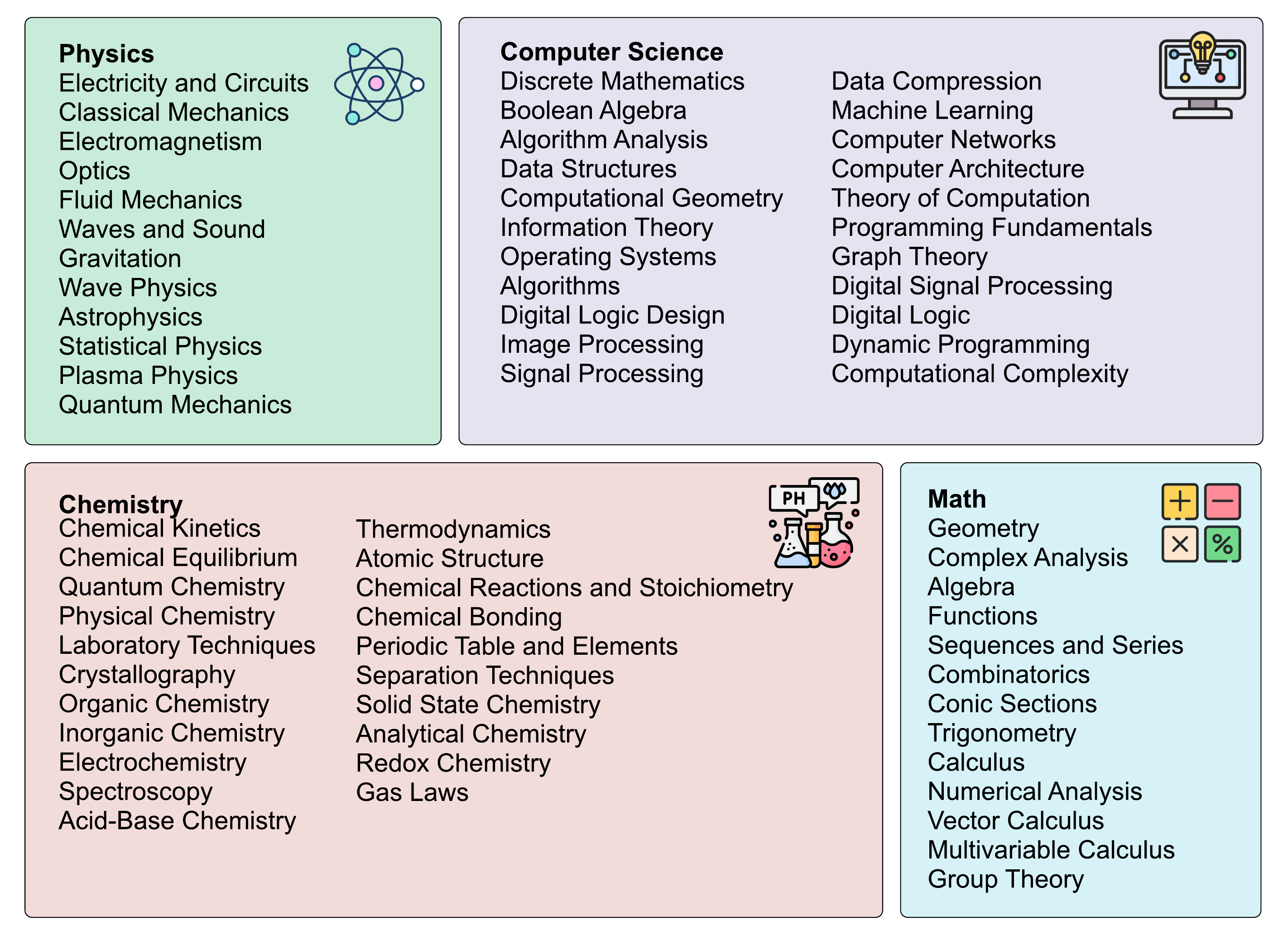}
    \caption{Subfields of TheoremExplainBench under Computer Science, Chemistry, Mathematics, and Physics.}
    \label{fig:dataset_main}
\end{figure*}

\subsection{LLM in Theorems Understanding}
LLMs have demonstrated remarkable capabilities in solving complex mathematical problems, including formal theorem proving and symbolic reasoning. To evaluate these abilities, researchers have introduced multiple benchmark datasets, primarily consisting of multiple-choice and short-answer question answering (QA) tasks~\cite{zhang2024multiplechoicequestionsefficientrobust, amini2019mathqainterpretablemathword, hendrycksmath2021}. Early studies centered on elementary to high school-level mathematics, leading to datasets such as Math23K~\cite{zhou2023learninganalogydiversequestions}, GSM8K~\cite{cobbe2021trainingverifierssolvemath}, and GeoQA~\cite{chen2022geoqageometricquestionanswering}. As LLM capabilities advanced, more domain-specific benchmarks emerged, extending evaluation to fields like science reasoning (ScienceQA)~\cite{lu2022learn}, financial reasoning (FinQA)~\cite{chen2022finqadatasetnumericalreasoning}, and theorem comprehension (TheoremQA)~\cite{chen2023theoremqa}. These datasets collectively assess LLMs' ability to solve mathematical and scientific problems up to the university level. However, existing benchmarks remain predominantly text-based, overlooking the role of visual intuition in mathematical reasoning. Many mathematical concepts are best understood through structured diagrams and dynamic representations, which current LLM evaluations fail to capture. To address this gap, we introduce an AI framework to generate theorem explanations in long-form videos, integrating symbolic derivations with structured visualizations to enhance comprehension.

\subsection{LLM in Visualizations}
Recent advancements in AI-driven visualization have enabled AI systems to generate structured visual content from textual descriptions~\cite{li2024visualizationgenerationlargelanguage}. These models typically process text-based inputs and produce programmatic representations, which are then converted into visual outputs~\cite{ritchie2023neurosymbolicmodelscomputergraphics, goswami2025plotgenmultiagentllmbasedscientific}. This approach has been applied across various domains, including scientific visualization~\cite{yang2024matplotagent}, data representation~\cite{galimzyanov2024drawingpandasbenchmarkllms}, and motion graphics~\cite{zhang2023motiongraphics}. Efforts such as Drawing-Pandas~\cite{galimzyanov2024drawingpandasbenchmarkllms} have introduced benchmarks for evaluating code-based plotting in Matplotlib and Seaborn. Follow-up works like MatPlotAgent\cite{yang2024matplotagent} demonstrated that agentic approaches outperform agentless methods in visualization generation, while PlotGen~\cite{goswami2025plotgenmultiagentllmbasedscientific} incorporated multimodal feedback for iterative refinement, further improving visualization quality. Our work is the first to explore AI-driven visualization for generating animated theorem explanations, seamlessly integrating step-by-step symbolic derivations with structured motion graphics, bridging the gap between mathematical reasoning and visual comprehension.

\section{Method}

\subsection{Task Definition}

\textbf{Model Input.} The model receives a theorem along with a short description that provides context, which helps the model identify the theorem.

\noindent\textbf{Model Output.} The model is to output a video that combines animations, structured derivations, and voiceover narration to provide a multimodal and comprehensive explanation of the theorem. The video is expected to be longer than a minute, featuring long animations across different scenes, with narration guiding the viewer through step-by-step proofs and real-world applications.

\begin{table*}[!t]
\centering
\scalebox{0.85}{
\begin{tabular}{l|ccc|cccc|c}
\toprule
Agent & Easy & Medium & Hard & Math & Phys & CS & Chem & Overall\\
\midrule
GPT-4o & 61.3\% & 57.5\% & 46.2\% & 61.7\% & 55.0\% & 58.3\% & 45.0\% & 55.0\%\\
GPT-4o + RAG & 42.5\% & 57.5\% & 37.5\% & 70.0\% & 40.0\% & 41.7\% & 31.7\% & 45.8\%\\
Claude 3.5-Sonnet v1 & 2.5\% & 1.2\% & 2.5\% & 1.7\% & 1.7\% & 1.7\% & 3.3\% & 2.1\% \\
Claude 3.5-Sonnet v1 + RAG & 18.8\% & 13.8\% & 11.2\% & 23.3\% & 10.0\% & 20.0\% & 5.0\% & 14.6\%\\
Gemini 2.0-Flash & 20.0\% & 11.2\% & 12.5\% & 16.7\% & 8.3\% & 21.7\% & 11.7\% & 14.6\%\\
Gemini 2.0-Flash + RAG & 23.8\% & 21.2\% & 16.2\% & 26.7\% & 15.0\% & 20.0\% & 20.0\% & 20.4\%\\
o3-mini (medium) & \textbf{93.8\%} & \textbf{91.2\%} & \textbf{96.2\%} & \textbf{95.0\%} & \textbf{93.3\%} & \textbf{93.3\%} & \textbf{93.3\%} & \textbf{93.8\%} \\
o3-mini (medium) + RAG & 83.8\% & 82.5\% & 80.0\% & 81.7\% & 90.0\% & 88.3\% & 68.3\% & 82.1\%\\
\bottomrule
\end{tabular} }
\caption{Agent success rate in generating complete videos across different difficulty levels and subjects.}
\label{tab:succ_diff_subject}
\end{table*}

\begin{table*}[!t]
\centering
\scalebox{0.85}{
\begin{tabular}{l|ccccc|c}
\toprule
Agent & Accuracy & Visual & Logical & Element & Visual & Overall \\
  & and Depth & Relevance & Flow & Layout & Consistency & Score \\
\midrule
GPT-4o & 0.79 &  0.79 &  \textbf{0.89} &  0.59 &  0.87 & 0.78 \\
GPT-4o + RAG  &  0.75 &  0.77 &  0.88 &  0.57 &  0.86 &  0.76\\
Claude 3.5-Sonnet v1 &  0.75 &  \textbf{0.87} &  0.88 &  0.57 &  \textbf{0.92} &  \textbf{0.79}\\
Claude 3.5-Sonnet v1 + RAG&  0.67 &  0.79 &  0.69 &  0.65 &  0.87 &  0.71\\
Gemini 2.0 Flash  &  \textbf{0.82} &  0.77 &  0.80 &  0.57 &  0.88 &  0.76\\
Gemini 2.0 Flash + RAG  &  0.79 &  0.75 &  0.84 &  0.58 &  0.87 &  0.76\\
o3-mini (medium) &  0.76 & 0.76 &  \textbf{0.89} &  0.61 &  0.88 &  0.77 \\
o3-mini (medium) + RAG &  0.75 &  0.75 &  0.88 &  0.61 &  0.88 &  0.76\\
\midrule
Human-made Manim Videos &  0.80 &  0.81 &  0.70 &  \textbf{0.73} &  0.87 &  0.77\\
\bottomrule
\end{tabular} }
\caption{Performance of our proposed metrics on successfully generated long-form videos by the agents.}
\label{tab:overall_performances}
\end{table*}

\subsection{TheoremExplainAgent (TEA)}

We develop TheoremExplainAgent (TEA), an agentic pipeline designed to automate the generation of videos using multiple specialized agents as shown in Figure~\ref{fig:method_main}. The process begins with the planner agent, which creates a high-level video plan according to the specified theorem. This plan consists of multiple scenes, each corresponding to a key segment of the resulting video. Once the initial plan is created, the planner agent refines the details of each scene, breaking them down into smaller components that define the specific visual elements, animations, and transitions needed. These detailed scene descriptions are then passed to the coding agent, which generates the corresponding Python code. The voiceover is also generated through a text-to-speech service. Finally, the Python scripts are executed to produce the final video, which reflects the narrative or instructional goals outlined in the video plan. If the generated Python code encounters an error, the coding agent will review the error and generate a revised version of the code. We set a maximum of $N$ attempts where $N = 5$. If this limit is exceeded, we mark the generation as unsuccessful. We found that at $N=0$, the success rate is extremely low while $N=5$ achieves up to 90\% success rate, as discussed in Table~\ref{tab:n_attempt_success}.

\textbf{Coding Toolkit.} We choose Manim~\cite{The_Manim_Community_Developers_Manim_Mathematical_2024} as the coding toolkit because it is a popular open-source Python library designed for creating mathematical animations and educational videos through code-driven visualizations. YouTube channels such as 3Blue1Brown~\cite{3b1bvideo} have demonstrated how Manim-made videos can convey complex mathematical concepts in an intuitive way. In our context, the coding agent translates each scene's specifications into executable Manim scripts, which define objects such as text, shapes, graphs, or equations, along with their corresponding animations, timings, and transitions.

\textbf{Agentic Retrieval-Augmented Generation.} To enhance code generation ability, we implemented a multifaceted retrieval-augmented generation (RAG) approach, leveraging the Manim documentation as the primary knowledge base. Unlike a single monolithic retrieval step, our agentic approach first classifies whether the theorems are suitable for using specific Manim plugins. Then it generates relevant queries at different stages of the video creation process: (1) during storyboard generation, to retrieve visual examples and related concepts; (2) during technical implementation, to fetch specific code snippets and usage patterns; and (3) during error correction, to diagnose issues and suggest solutions. These queries are cached to prevent redundant computations, and the agent dynamically selects the most relevant documents based on a relevance scoring threshold, ensuring efficient and precise retrieval.

\begin{table*}[!t]
\centering
\scalebox{0.9}{
\begin{tabular}{lcccccc}
\toprule
Category & $N=0$ & $N=1$ & $N=2$ & $N=3$ & $N=4$ & $N=5$ \\
\midrule
Difficulty: Easy & 7\% / 5\% & 51\% / 33\% & 73\% / 66\% & 86\% / 71\% & 91\% / 73\% & \textbf{93\%} / \textbf{73\%} \\
Difficulty: Medium & 0\% / 5\% & 33\% / 43\% & 75\% / 66\% & 83\% / 72\% & 88\% / 76\% & \textbf{91\%} / \textbf{77\%} \\
Difficulty: Hard & 3\% / 3\% & 46\% / 35\% & 81\% / 51\% & 90\% / 68\% & 95\% / 71\% & \textbf{96\%} / \textbf{73\%} \\
\bottomrule
\end{tabular}}
\caption{Combined cumulative theorem success rates (Baseline / RAG) for o3-mini (medium) with varying N attempts. The overall success rate significantly improves with more attempts, reaching its peak at $N=5$.}
\label{tab:n_attempt_success}
\end{table*}

\subsection{TheoremExplainBench (TEB)}

We curate an evaluation dataset comprising 240 theorems from various disciplines, including Computer Science, Chemistry, Mathematics, and Physics. Each entry includes the theorem name and a contextual description, sourced from OpenStax~\cite{openstax} and LibreTexts~\cite{libretexts}. To facilitate structured assessment, the theorems are categorized into three difficulty levels: Easy (high school level), Medium (undergraduate level), and Hard (graduate level), with 80 entries in each category. TheoremExplainBench (TEB) features 68 sub-fields that cover a wide range of domains as shown in Figure~\ref{fig:dataset_main}.

To fully define this novel problem, we propose a comprehensive evaluation metric applicable to both human-created and AI-generated explanatory videos, ensuring a standardized assessment across different content sources. Our metric evaluates videos across five key dimensions. The first three dimensions assess the factual correctness of explanations, while the last two dimensions evaluate the perceptual quality of the videos.

\textbf{Accuracy and Depth.} Evaluates whether the narration provides a precise and well-structured explanation of the theorem, offering both intuitive insights and rigorous justifications for why it holds.

\textbf{Visual Relevance.} Assesses whether the video frames effectively align with the theorem's concepts and derivations, reinforcing the explanation through appropriate visual representations.

\textbf{Logical Flow.} Examines whether the video follows a clear and coherent structure, ensuring a logical progression that builds upon ideas effectively.

\textbf{Element Layout.} Evaluates whether visual elements are well-positioned and appropriately sized within the frame, avoiding unintended overlap and ensuring clarity in presentation.

\textbf{Visual Consistency.} Assesses whether the motions are smooth, and whether the visual style remains uniform across frames.

In our metric implementation, Accuracy \& Depth and Logical Flow are assessed using text-based evaluation with GPT-4o~\cite{openai2023gpt4}. The text elements are extracted from video transcripts in SubRip (SRT) format. For Visual Relevance and Element Layout, we apply image processing techniques to identify key frames and use GPT-4o to assign scores for each dimension. To evaluate motions in Visual Consistency, we utilize Gemini 2.0-Flash~\cite{deepmind2025gemini2flash} to analyze chunked video segments. The overall score (ranging from 0 to 1) is then computed as the geometric mean of all dimensions. To ensure output stability, we employ greedy decoding (i.e., temperature $= 0$) in the LLM evaluations.

To validate the effectiveness of our evaluation metrics, we conducted a small-scale human study. We sampled 40 videos from our results, selecting 10 from each discipline in TheoremExplainBench. We then recruited 12 experienced STEM student annotators to participate in the study. The rating process followed the same five evaluation dimensions as our proposed metrics, with human raters selecting scores from [0, 0.5, 1]. To assess alignment between our metrics and human evaluations, we computed the Spearman correlation on the sampled subset. To ensure result reliability, we measured inter-rater agreement of 3 people using Krippendorff's alpha~\cite{Krippendorff2011ComputingKA}, which is more suitable than Fleiss' Kappa~\cite{fleiss1973equivalence} due to the ordinal nature of the ratings. Additionally, to contextualize human performance, we sourced 10 human-made theorem explanation videos from YouTube for comparison.

\section{Experimental Results}

For the agent candidates in TheoremExplainAgent, we experimented with GPT-4o~\cite{openai2023gpt4}, Gemini 2.0 Flash~\cite{deepmind2025gemini2flash}, Claude 3.5 v1~\cite{anthropic2025claude3}, and o3-mini~\cite{openai2025o3mini}. Each candidate was used for both the planner agent and coding agent, ensuring consistency across configurations. We evaluated all agents across 240 theorems from TheoremExplainBench, comparing their performance under different setups. Our findings indicate that an agentless approach fails to generate videos longer than 20 seconds, whereas TheoremExplainAgent successfully produces videos of up to 10 minutes. Consequently, all experimental results presented in this paper are based on the agentic approach.

Table~\ref{tab:succ_diff_subject} reveals that the
success rate in generating long-form theorem explanation videos varies significantly across difficulty levels and subjects. Overall, o3-mini consistently outperforms other models, maintaining high success rates across both easy and hard tasks, as well as across different STEM domains.  In contrast, GPT-4o performs moderately well but show a declining success rate as complexity increases, suggesting difficulties in handling longer and more structured explanations. Gemini 2.0-Flash struggles the most, with notably lower success rates across all conditions. Across subjects, Mathematics tends to have the highest success rates, whereas Chemistry appear to be the most challenging domain. This observation may be attributed to the fact that complex objects in Chemistry, such as flask shapes and atoms, are more challenging to illustrate than simpler primitives in Mathematics, like triangles.

\begin{table}[!t]
\centering
\scalebox{0.85}{
\begin{tabular}{l|cc}
& Spearman & Krippendorff's $\alpha$ \\
\toprule
Accuracy and Depth & 0.14 & 0.45 \\
Visual Relevance & \textbf{0.72} & 0.36 \\
Logical Flow & 0.16 & 0.56 \\
Element Layout & \textbf{0.42} & 0.31 \\
Visual Consistency & 0.17 & 0.36 \\
\bottomrule
\end{tabular} }
\caption{Correlation on Metric-Human correlation (Spearman) and Inter-rater Agreement (Krippendorff's alpha) for the five evaluation dimensions.}
\label{tab:human_corr}
\end{table}

Given the successfully generated videos, we compiled Table~\ref{tab:overall_performances} to present the metric results. Among the evaluated models, GPT-4o and o3-mini performed the best overall, both achieving strong scores across multiple dimensions. GPT-4o excelled in accuracy and depth, as well as logical flow, while o3-mini demonstrated the strongest performance in logical flow and a solid element layout. On the other hand, Gemini 2.0 Flash with RAG performed the weakest overall, struggling particularly with element layout and logical flow, indicating challenges in maintaining structured and visually coherent outputs. Human-made Manim videos, while scoring the similar overall among AI-generated results, achieved the highest visual relevance and element layout. This may be because AI-generated videos tend to exhibit minor issues like overlapping elements and misalignment, which can affect clarity and structure. Interestingly, human-made videos scored lower in logical flow. This may be due to the more natural and less structured narration in human explanations, which often prioritize engagement over strict logical progression. In contrast, AI-generated videos tend to maintain a consistent logical structure, adhering closely to predefined formats. However, this rigidity may sometimes come at the cost of expressiveness and contextual adaptability, making human explanations feel more fluid and accessible despite their lower scores in formal evaluation metrics.

Our experiments with the RAG setup yielded mixed results, as shown in Table~\ref{tab:succ_diff_subject} and Table~\ref{tab:overall_performances}. While RAG was intended to enhance function understanding and streamline object construction, its effectiveness proved inconsistent. Although retrieval of documentation and code examples provided additional context, the results often misaligned with specific use cases. Many retrieved references were overly generic or lacked relevance, leading to incorrect function calls and suboptimal parameter choices. These findings are consistent with previous research highlighting the critical importance of retrieval quality. Poorly structured documentation and imprecise retrieval can significantly compromise the effectiveness of RAG-based approaches~\cite{soman2024observationsbuildingragsystems}.

We also conducted an ablation study on the cumulative success rates of our o3-mini (medium) models by varying the retries value N $\in$ \{0, 1, 2, 3, 4, 5\}. The ``N-attempt success rate" is defined as the percentage of theorems for which all constituent scenes are successfully rendered. This metric evaluates the cumulative performance as we allow for more attempts to correct initial errors. Both the baseline and RAG-enhanced o3-mini models demonstrate a substantial increase in theorem success rates as N increases, indicating that multiple attempts effectively mitigate the impact of initial code generation failures. The detailed results are presented in Table~\ref{tab:n_attempt_success}.

\begin{figure*}[!t]
    \centering
    \includegraphics[width=0.85\linewidth]{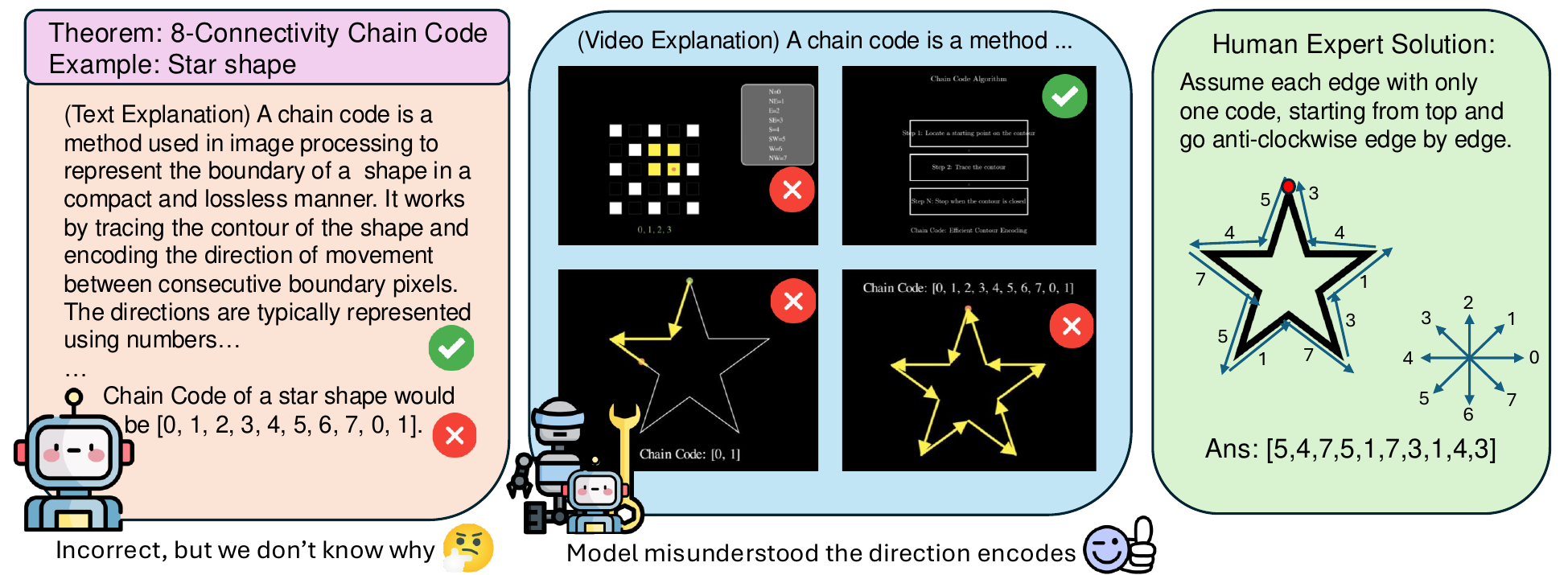}
    \caption{Visualizations expose reasoning errors more clearly than text, making it easier to diagnose model mistakes.}
    \label{fig:casestudy_2_main}
\end{figure*}

\subsection{Correlation Study}

From Table~\ref{tab:human_corr}, we observe that our proposed metrics show strong alignment with human ratings in Visual Relevance and Element Layout, while demonstrating weaker correlations in Accuracy \& Depth, Logical Flow, and Visual Consistency. This suggests that humans are particularly sensitive to visual aspects, such as spatial layouts, but may struggle with evaluating long-form text or audio-based content in detail. Visual Consistency appears to be more subjective, which may explain its relatively lower correlation with human ratings. Additionally, Accuracy \& Depth and Logical Flow exhibits the weakest correlation with human judgments, likely due to differences in how LLM and humans assess coherence. Humans can tolerate informal flow, while LLMs may penalize it. On the other hand, human ratings across all dimensions show moderate inter-rater agreement, as indicated by Krippendorff's alpha values. Notably, text-based dimensions achieve slightly higher agreement than visual-based ones, suggesting that textual evaluations are more consistently interpreted among raters.

\subsection{Interpretability Study}
\label{sec:supp_interpretability_study}

We found that visual explanations more effectively reveal reasoning errors than text, facilitating error diagnosis. From Figure~\ref{fig:casestudy_2_main}, we observe that while the text-based explanation allows us to detect that the model's answer is incorrect, it does not provide insight into why the mistake occurred. It seems the model understand the chain code theorem, but it applies it incorrectly. Such explanation is making it difficult to pinpoint the exact reasoning flaw. In contrast, the video-based explanation clearly exposes the model's misunderstanding, as incorrect movement direction encodes and misplaced arrows reveal how the model misinterpreted the chain coding process. This demonstrates that visual explanations not only confirm incorrect reasoning but also uncover the underlying cause of errors, making them a more effective diagnostic tool for analyzing AI-generated outputs.

To quantitatively assess whether video explanations better highlight LLM reasoning problems, we designed a human study. 15 participants were first shown a textual explanation of a theorem containing a subtle reasoning flaw and asked to judge its correctness. They were then shown a video explanation of the same theorem, embodying the same reasoning flaw, and asked to re-evaluate. Participants also rated the intuitiveness of both explanations on a scale of 1 to 5. Initially, all 15 participants judged the textual explanation as correct. After watching the video, 9 participants (60\%) revised their judgment to "incorrect," successfully identifying the conceptual flaw that was only apparent through the visual narration. The average intuitiveness rating improved from 3.3 for the textual explanation to 3.9 for the video explanation. This suggests that multimodal explanations can be more effective in revealing and helping users understand reasoning errors.

\subsection{Error Analysis}
\label{sec:error_analysis}

We analyzed the error logs from unsuccessful runs in the TheoremExplainAgent video generation process and identified three primary failure categories. The most common issue was Manim code hallucinations, which accounted for the majority of failures. These errors involved nonexistent functions, modules, object properties, or image assets, as well as incorrect function signatures with invalid parameter types and numbers, reflecting a misunderstanding of the Manim API. The second major issue stemmed from LaTeX rendering errors, primarily due to syntax mistakes and improper handling of special characters in mathematical expressions. Lastly, general coding errors were observed, including missing imports, undefined variables, and computational mistakes in NumPy-based operations. These findings reveal key challenges across LLMs, underscoring the need for better code reliability and API understanding in AI-generated videos.

\begin{figure*}[!t]
    \centering
    \includegraphics[width=1.0\linewidth]{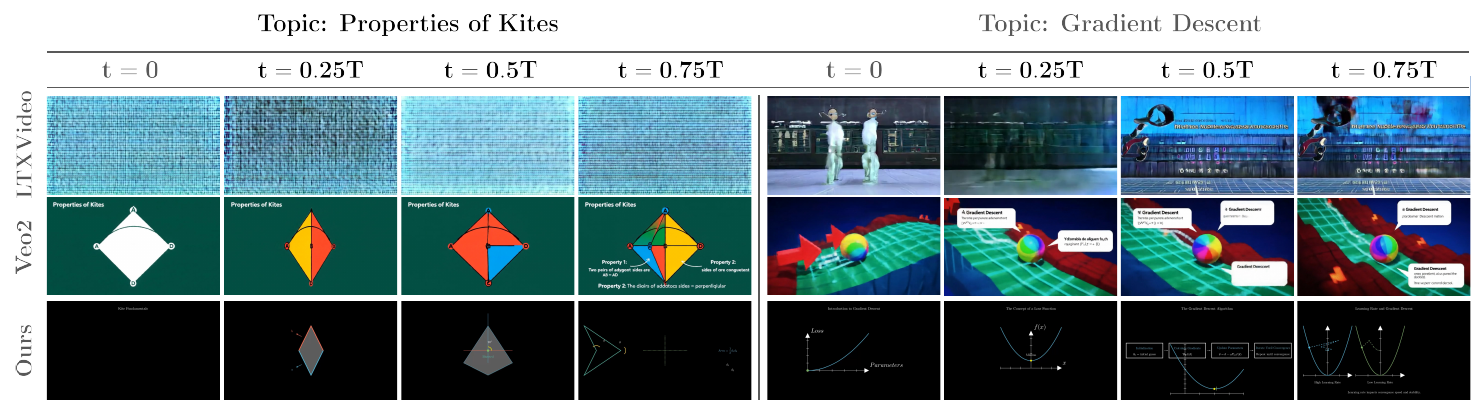}
    \caption{Video quality comparison by video frame between LTXVideo, Veo2 and TheoremExplainAgent, where $t$ represents the proportion of the total video duration. $t$ correspond to the time. The approximate number of $T$ for LTXVideo, Veo2 and TheoremExplainAgent are 7 seconds, 8 seconds, and 5 minutes respectively.}
    \label{fig:text-to-video-comparison}
\end{figure*}

\subsection{Case Study}
\label{sec:case_study}

We included representative video outputs in Figure~\ref{fig:gallery}, demonstrating that TheoremExplainAgent is capable of generating high-quality exploratory videos. For example, in Mathematics, the model effectively visualizes concepts such as Riemann sums, using animated grids and function plots to illustrate integral approximations. In Chemistry, the system successfully explains the Octet Rule, leveraging atomic models to depict electron sharing and bonding interactions. In Physics, it generates electromagnetic wave simulations, showcasing wave propagation and spectral analysis. In Computer Science, it produces a clear demonstration of Run-Length Encoding, using side-by-side comparisons of raw and compressed data representations. Videos in Mathematics, Physics, and Computer Science typically show higher visual quality and coherence than those in Chemistry. One notable observation is that Chemistry-related visualizations often rely on simple geometric primitives to illustrate complex lab equipment and molecules, which can limit their clarity and effectiveness. Additionally, most of the generated videos exhibit minor element layout issues, such as overlapping texts, inconsistent sizes, or suboptimal object positioning, which slightly affects the overall presentation quality, as illustrated in Figure~\ref{fig:fail_gallery}.

\subsection{Text-to-Video Model Baselines}

To access whether explicit reasoning capabilities provided by LLM-based agents are essential for generating coherent theorem explanations, we conducted a baseline comparison using non-LLM-based text-to-video models. Specifically, we examined the performance of the recent open-source model LTXVideo~\cite{HaCohen2024LTXVideo} and the closed-source model Veo2~\cite{veo2}. For this comparison, we prompted each model with: ``a Manim-style explanatory video explaining <theorem>'' for 20 randomly selected theorems from TheoremExplainBench. As illustrated in Figure~\ref{fig:text-to-video-comparison}, the resulting videos were frequently visually incoherent, often manifesting as random noise or lacking meaningful relation to the intended scientific content. These outputs also lacked the structured, pedagogical qualities necessary for effective theorem explanations. Moreover, these models do not include voiceover narration, an integral component of our multimodal output. This highlights that while text-to-video models can synthesize visual content from text, they lack the explicit reasoning capabilities required to generate domain-specific explanatory videos. In contrast, TheoremExplainAgent enables the creation of detailed, pedagogically sound videos that align with the theorems.

\section{Conclusion}
This paper introduces TheoremExplainAgent, a novel agentic approach for generating multimodal theorem explanations through structured video content. We demonstrates that integrating visual explanations significantly enhances the clarity and interpretability of theorem reasoning, surpassing text-based methods alone. We also present a benchmark spanning multiple disciplines with five automated evaluation metrics. Experiments reveal that agentic planning is crucial for producing long-form, coherent explanations, with o3-mini achieving the highest success rate and overall performance. However, challenges remain in visual element layout, emphasizing the need for improved spatial reasoning and refinement in AI-generated animations. Additionally, our findings underscore the importance of multimodal explanations in identifying reasoning flaws that text-based assessments often miss, reinforcing the role of structured visual communication in AI-driven theorem understanding.

\clearpage
\newpage

\section{Limitations}

While our approach demonstrates the potential of AI-generated multimodal theorem explanations, several limitations remain.

\paragraph{Visual Structuring Challenges.}
TheoremExplainAgent exhibits limitations in visual structuring, with issues such as misaligned text, overlapping shapes, and inconsistent sizes. These visual imperfections, though sometimes subtle, can distract from the intended educational value and hinder comprehension, particularly in complex topics like Chemistry. As mentioned in Section~\ref{sec:case_study} and illustrated in Table~\ref{tab:overall_performances}, both human and automated evaluations identified shortcomings in element layout and visual consistency, highlighting the need for further refinement in visual design.

\paragraph{Inconsistent Retrieval-Augmented Generation.}
The reliance on retrieval-augmented generation (RAG) to support Manim code generation proved inconsistent in practice. As noted in Section~\ref{sec:case_study} and shown in Table~\ref{tab:cost}, retrieved code snippets were often irrelevant or overly generic, resulting in hallucinated or incorrect function calls and suboptimal parameter choices. Moreover, the use of RAG significantly increased token usage and inference time, raising concerns about scalability and cost efficiency.

\paragraph{Evaluation Metric Limitations.}
Our automated metrics for Accuracy \& Depth (Spearman $\rho=0.14$), Logical Flow ($\rho=0.16$), and Visual Consistency ($\rho=0.17$) show weak correlation with human ratings, while Visual Relevance ($\rho=0.72$) and Element Layout ($\rho=0.42$) achieve moderate to high agreement with good p-values (0.001 and 0.03 respectively) to show statistical significance. We acknowledge that our automated metrics have room for improvement, particularly for aspects like narrative coherence and the nuances of visual consistency. Current video understanding capabilities in Vision Language Models (VLMs) are still in early stages of development. Future work could explore fine-tuning specialized video understanding models (e.g., VideoScore~\cite{he2024videoscorebuildingautomaticmetrics}) for assessing these types of tasks more accurately. We agree that visual consistency is part of the visual assessment and will refine the evaluation design in future iterations.

\paragraph{Code Generation Fragility.}
The system remains vulnerable to common coding errors, including LaTeX rendering failures, incorrect function calls, and general Python errors such as undefined variables and missing imports. These issues were detailed in Section~\ref{sec:error_analysis} and highlight the fragile nature of current code-generation approaches. Despite the implementation of a retry mechanism with up to five attempts, code reliability remain critical challenges.

\section{Potential Risks}

AI-generated explanations have the potential to mislead users if errors go undetected, leading to false confidence in incorrect reasoning. This poses a risk where unverified AI-generated content could propagate misconceptions or misinformation if widely disseminated without proper validation. Ensuring the accuracy and reliability of AI-generated explanations remains a critical challenge.

\section{Artifacts}
We experimented TheoremExplainAgent with GPT-4o~\cite{openai2023gpt4}, Gemini 2.0 Flash~\cite{deepmind2025gemini2flash}, Claude 3.5 v1 ~\cite{anthropic2025claude3}, and o3-mini~\cite{openai2025o3mini}. We are releasing the TheoremExplainBench on Huggingface dataset with MIT licence. It features 240 theorems across Computer Science, Physics, Chemistry and Math subjects. 

\section{Computational Experiments}
All the experiments were conducted on a NVIDIA A100-SXM4-80GB GPU. 
Approximately 1500 US dollars were spent on API call for closed-source model experiments.

\section{Acknowledgement}
We express our gratitude to Votee AI for sponsoring API calls from closed-source models. We also thank Xueguang Ma, Dongfu Jiang, Zhi-Rui Tam, Chiu-Wai Yan, and Kelly Chiu for their insightful discussions. 

\bibliography{acl_latex}

\clearpage
\appendix

\begin{figure*}[!t]
    \centering
    \includegraphics[width=1.0\linewidth]{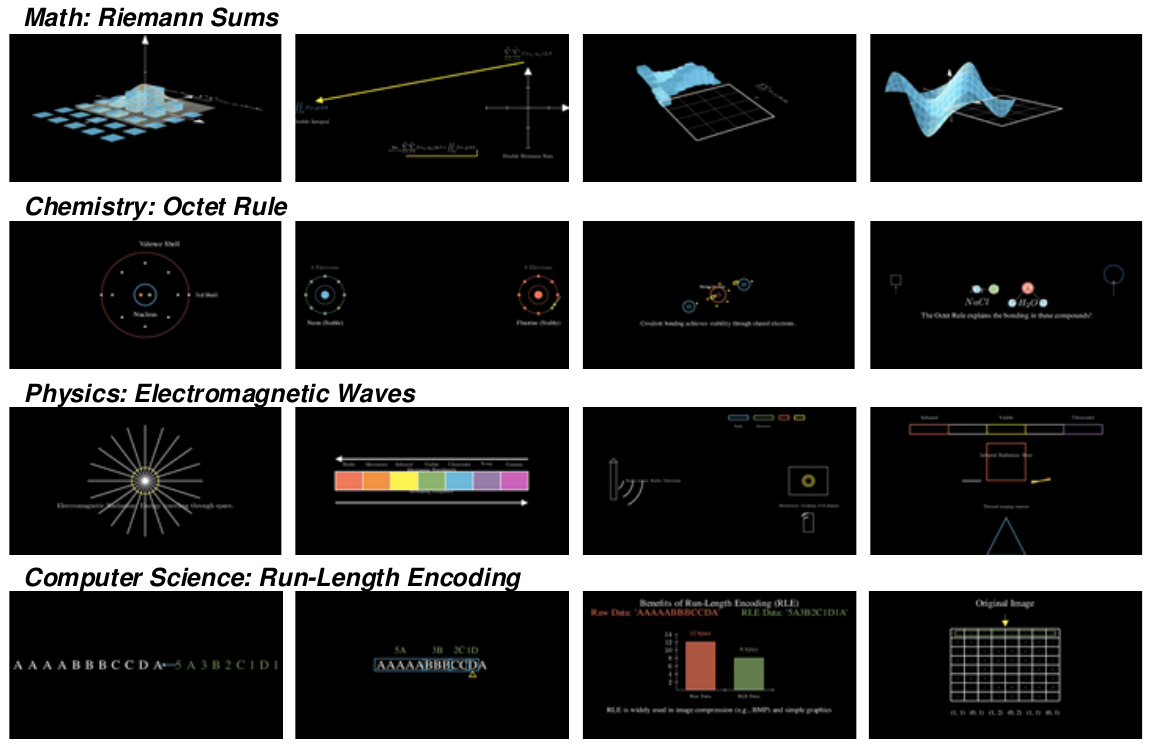}
    \caption{We show the high-quality videos generated by TheoremExplainAgent,across the four STEM domains.}
    \label{fig:gallery}
\end{figure*}

\begin{figure*}[!t]
    \centering
    \includegraphics[width=1.0\linewidth]{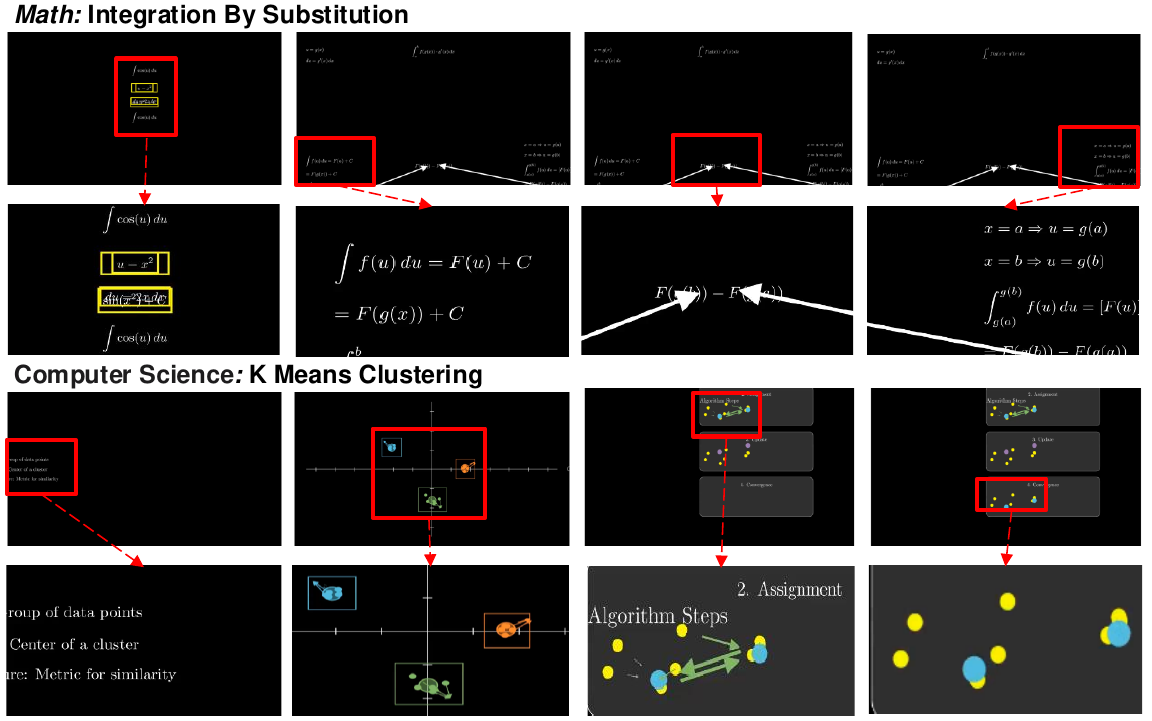}
    \caption{We show the poorly generated videos from TheoremExplainAgent, zooming in the artifacts.}
    \label{fig:fail_gallery}
\end{figure*}

\clearpage

\begin{figure*}[!t]
    \centering
    \includegraphics[width=1.0\linewidth]{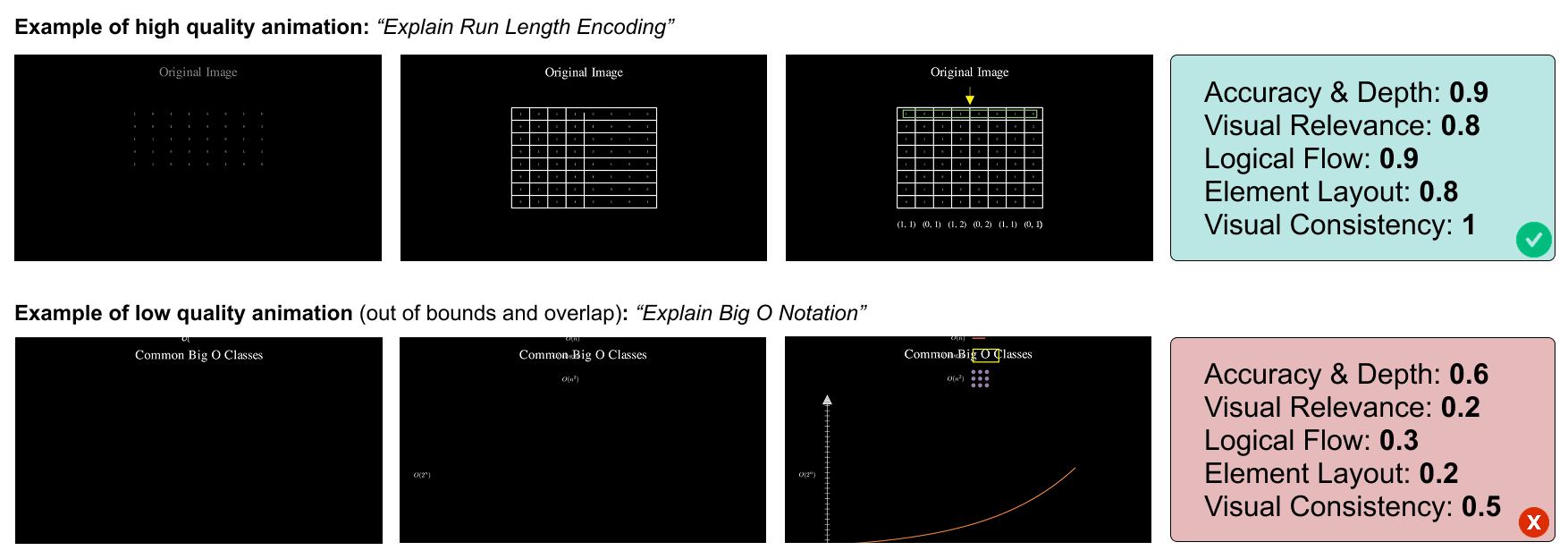}
    \caption{Comparison on a scene of a high quality animation and a low quality animation.}
    \label{fig:casestudy_1_main}
\end{figure*}

\begin{figure*}[!t]
    \centering
    \includegraphics[width=1.0\linewidth]{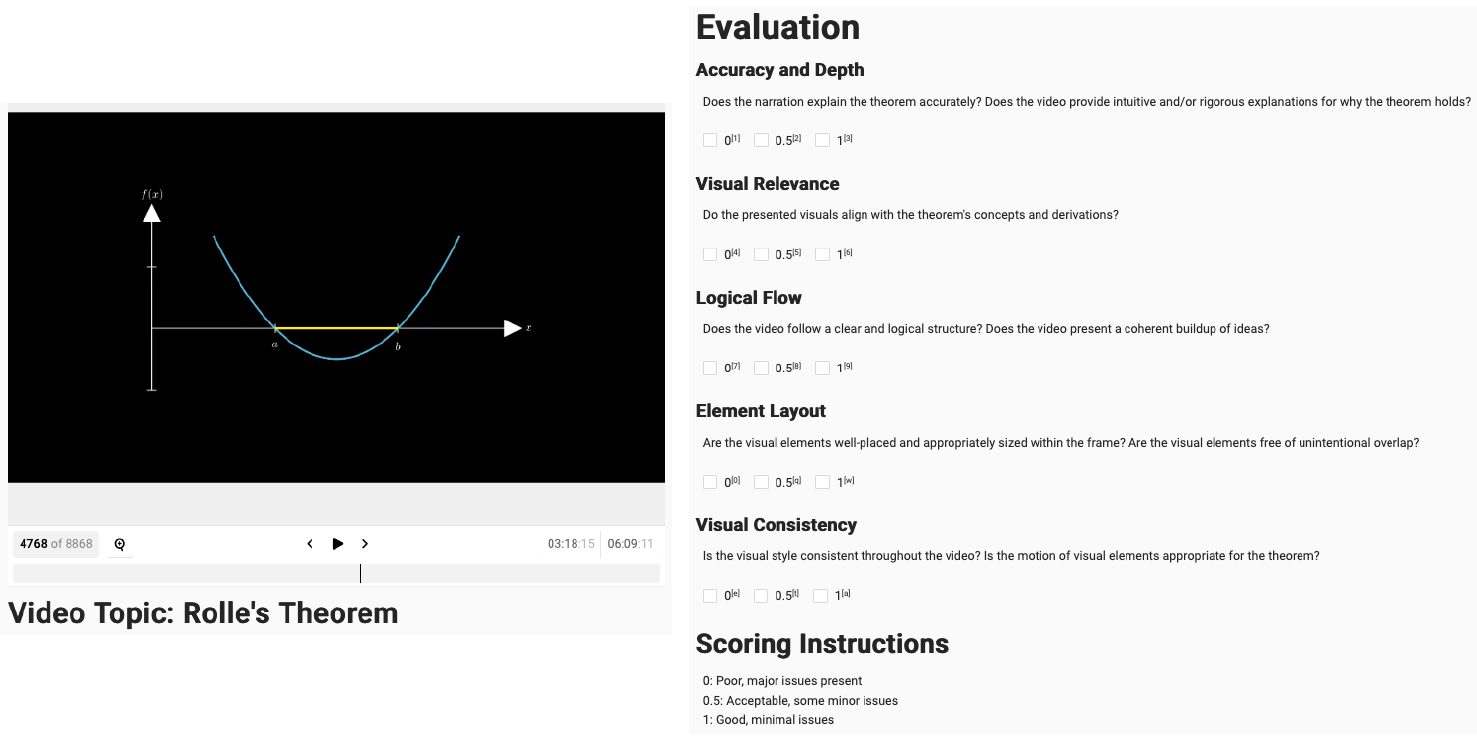}
    \caption{The user interface of our annoatation website.}
    \label{fig:labelstudio}
\end{figure*}

\clearpage

\section{Gallery}
\label{sec:gal}
In Figure~\ref{fig:gallery} we present the high-quality videos generated by TheoremExplainAgent across four STEM domains. The images are extracted from different scenes in the videos, showing the consistency of the topic. In Figure~\ref{fig:fail_gallery} we present the poorly generated videos from TheoremExplainAgent and examine their artifacts. In Figure~\ref{fig:casestudy_1_main} we compare a high quality animation and a low quality animation, and how they were rated with our proposed metric.




    


\section{Runtime Statistics}

We report the runtime and cost statistics in Table~\ref{tab:cost}, assuming 4 fixed codes and 7 scenes per video, we evaluate the cost, inference time, and latency of different language models, and find that the Claude 3.5-Sonnet v1 model has the longest inference time (2240-2380s), while the Gemini 2.0-Flash and GPT-4o are the fastest (around 1120s).The RAG integration increases the number of input tokens significantly. RAG integration significantly increases the number of input tokens, with Claude 3.5-Sonnet v1 + RAG being the most used (1,050,000). Output tokens are less variable, with the o3-mini model generating the most tokens (154,000). The Gemini 2.0-Flash model is the most cost-effective (\$0.10-\$0.16), while the Claude 3.5-Sonnet v1 + RAG is the most expensive (\$4.67).

\section{Preference Study}
To complement our automated metrics, we conducted a human evaluation study. 18 annotators were asked to rank videos on 3 topics generated by three different models (o3-mini, Gemini 2.0 Flash, and GPT-4o) for the same set of theorems, without revealing the model names. Annotators evaluated the videos based on clarity and visual appeal. The results, summarized in Table~\ref{tab:pairwise_eval}, indicate that videos generated by Gemini 2.0 Flash were most frequently ranked highest for both clarity and visual appeal. o3-mini generated videos were preferred over those from GPT-4o. This human preference study provides complementary insights into the perceived quality of the generated videos.

\section{Implementation Details}
\label{sec:supp_implementation_details}

\subsection{Human Annotation Process}
We recruited 12 student volunteers in our annotation process for the human metric. We explained to the annotators that their annotations were to be used in our study only and would not be released publicly.

We show the user interface of our annotation website in Figure~\ref{fig:labelstudio}, including the instructions presented to our annotators. We supplement each of the dimensions with guiding questions to clarify what the annotators should score.

\subsection{Technical Implementation}
To aid reproducibility, we provide the following key implementation details. \textbf{Manim Versions:}
\begin{itemize}
    \item Manim Community Edition version 0.18.1,
    \item ManimPango 0.6.0,
    \item manim-physics 0.4.0,
    \item manim-ml~\cite{helbling2023manimml} 0.0.24,
    \item manim-chemistry 0.4.4,
    \item manim-dsa~\cite{Missagia_Manim_DSA} 0.2.0, 
    \item manim-circuit 0.0.3
\end{itemize}

\textbf{RAG System:} We utilize ChromaDB as the vector store. Documentation (Markdown and Python files from Manim core and specified plugins) is chunked using Langchain's \texttt{RecursiveCharacterTextSplitter} with default settings for the respective languages. The embedding model used is \texttt{text-embedding-005} from Google Vertex AI.

\textbf{RAG Retrieval Threshold:} A relevance score threshold of 0.5 is used during similarity search in both core and plugin vector stores. We retrieve k=2 documents per query by default.

\subsection{Prompt Templates}
\label{sec:appendix_prompt}

\NewTColorBox{scenePlanBox}{ s O{!htbp} }{%
  floatplacement={#2},
  IfBooleanTF={#1}{float*,width=\textwidth}{float},
  colframe=gray!50!black,colback=gray!10!white,
  title=Scene Plan Generation Prompt Template,
  }
\NewTColorBox{codeGenerationBox}{ s O{!htbp} }{%
  floatplacement={#2},
  IfBooleanTF={#1}{float*,width=\textwidth}{float},
  colframe=blue!60!black, colback=blue!5!white,
  title=Code Generation Prompt Template,
  }
\NewTColorBox{codeFixingBox}{ s O{!htbp} }{%
  floatplacement={#2},
  IfBooleanTF={#1}{float*,width=\textwidth}{float},
  colframe=pink!50!black,colback=pink!10!white,
  title=Code Fixing Prompt Template
  }
\NewTColorBox{evalBox}{ s O{!htbp} }{%
  floatplacement={#2},
  IfBooleanTF={#1}{float*,width=\textwidth}{float},
  colframe=green!50!black, colback=green!10!white,
  title=Evaluation Prompt Template
  }

We adapt Chain-of-Thoughts (CoT)~\cite{wei2023chainofthoughtpromptingelicitsreasoning} and Program-of-Thoughts (PoT)~\cite{chen2023programthoughtspromptingdisentangling} when we design the prompt for TheoremExplainAgent. We present our prompts templates in the end of the Appendix.

\subsection{Computational Resources and Costs}
The primary LLM computations were performed using closed-source model APIs. The average number of tokens, cost, and inference time per video generation are detailed in Table~\ref{tab:cost}. Local Manim rendering does not require a GPU for the types of animations generated.

\section{Potentials for Future Research}

Recent community efforts~\cite{AIManimVideoGenerator, GatekeepAI, GenerativeManim} have explored AI-driven Manim-based video generation for educational purposes. However, no scientific studies have systematically evaluated the effectiveness and robustness of these approaches. Our work introduces a novel agentic framework for generating multimodal theorem explanations and demonstrates that AI-generated videos can achieve performance comparable to human-made content, although the robustness is still limited. Nevertheless, further research is needed to assess their impact on AI's reasoning capabilities, visualization quality, and learning outcomes. Future directions include establishing benchmarks for AI-generated educational videos (within EdTech), integrating interactive elements to enhance engagement (within HCI/Visualization), and refining evaluation metrics to assess LLMs' multimodal explanation abilities (within NLP).

\section{User Feedback on Educational Use}
We conducted a small-scale user study with 15 students and 8 teachers at universities to gather feedback on the educational clarity and engagement of AI-generated videos on easy topics. Participants rated videos on two topics (Bubble Sort and Hamming Distance) on a scale of 1 to 5 for clarity and engagement. As shown in Table~\ref{tab:user_feedback}, the mean scores for clarity and engagement were high. Over 90\% of participants expressed interest in using such videos in a classroom setting, although the majority also believed there is room for content improvement.

\section{Definition of Agentic}
Our use of the term ``agentic" follows the definition~\cite{DBLP:books/aw/RN2020}: ``An agent is anything that can be viewed as perceiving its environment through sensors and acting upon that environment through actuators." In the context of TheoremExplainAgent, the language model acts as an agent that perceives inputs (such as the theorem description and feedback from code execution errors) and acts upon its environment by generating Manim code to create video scenes.

\begin{table*}[!t]
\centering
\scalebox{1}{
\begin{tabular}{l|cccc}
Agent & Input Tokens & Output Tokens & Cost(USD) & Time(s) \\
\toprule
GPT-4o & 350000 & 84000 & 1.71 & 1120 \\
GPT-4o + RAG & 840000 & 84000 & 2.94 & 1260 \\
Claude 3.5-Sonnet v1 & 350000 & 91000 & 2.42 & 2240 \\
Claude 3.5-Sonnet v1 + RAG & 1050000 & 101500 & 4.67 & 2380 \\
Gemini 2.0-Flash & 595000 & 119000 & 0.1 & 1120 \\
Gemini 2.0-Flash + RAG & 1120000 & 119000 & 0.16 & 1260 \\
o3-mini (medium) & 434000 & 154000 & 1.16 & 1680 \\
o3-mini (medium) + RAG & 945000 & 154000 & 1.72 & 1820 \\
\bottomrule
\end{tabular} }
\caption{Average output tokens, cost, and inference time for TheoremExplainAgent generating one full video.}
\label{tab:cost}
\end{table*}

\begin{table*}[!t]
\centering
\scalebox{0.9}{
\begin{tabular}{lcc}
\toprule
Model Agent & Clarity Top-Rank \% & Visual Top-Rank \% \\
\midrule
Gemini-2.0-Flash~\cite{geminiteam2024gemini} & \textbf{70.6\%} & \textbf{61.8\%} \\
o3-mini~\cite{openai2025o3mini} & 20.6\% & 23.5\% \\
GPT-4o~\cite{openai2023gpt4} & 8.8\% & 14.7\% \\
\bottomrule
\end{tabular}}
\caption{Preference Study: Percentage of times each model was ranked highest by human annotators for video clarity and visual appeal.}
\label{tab:pairwise_eval}
\end{table*}

\begin{table*}[!t]
\centering
\scalebox{0.8}{
\begin{tabular}{lcc}
\toprule
Metric & Topic 1: Bubble Sort & Topic 2: Hamming Distance \\
\midrule
Clarity (Mean $\pm$ SD) & 4.13 $\pm$ 0.87 & 4.35 $\pm$ 0.83 \\
Engagement (Mean $\pm$ SD) & 3.57 $\pm$ 1.16 & 3.78 $\pm$ 1.09 \\
Would recommend for classroom use & 13\% & 35\% \\
Would recommend for classroom use if video content improved & 87\% & 61\% \\
Would not recommend for classroom use & 0\% & 4\% \\
\bottomrule
\end{tabular}}
\caption{User Feedback on Educational Usefulness.}
\label{tab:user_feedback}
\end{table*}

\begin{scenePlanBox}*[!ht]
You are an expert in video production, instructional design, and \{topic\}. Please design a high-quality video to provide in-depth explanation on \{topic\}.

\vspace{10pt}
\textbf{Video Overview:}

Topic: \{topic\} \\
Description: \{description\}

\vspace{10pt}
\textbf{Scene Breakdown:}

Plan individual scenes. For each scene please provide the following:

\begin{itemize}
    \item Scene Title: Short, descriptive title (2-5 words).
    \item Scene Purpose: Objective of this scene. How does it connect to previous scenes?
    \item Scene Description: Detailed description of scene content.
    \item Scene Layout: Detailed description of the spatial layout concept. Consider safe area margins and minimum spacing between objects.
\end{itemize}

Please generate the scene plan for the video in the following format: ...
\end{scenePlanBox}

\begin{codeGenerationBox}*[!ht]
You are an expert Manim (Community Edition) developer. Generate executable Manim code implementing animations as specified, strictly adhering to the provided Manim documentation context, technical implementation plan, animation and narration plan, and all defined spatial constraints.

\vspace{10pt}
Think of reusable animation components for a clean, modular, and maintainable library, prioritizing code structure and best practices as demonstrated in the Manim documentation context. Throughout code generation, rigorously validate all spatial positioning and animations against the defined safe area margins and minimum spacing constraints. If any potential constraint violation is detected, generate a comment in the code highlighting the issue for manual review and correction.

\vspace{10pt}
\textbf{Input Context:} \\
...

\vspace{10pt}
\textbf{Code Generation Guidelines:} \\
...
\end{codeGenerationBox}

\begin{codeFixingBox}*[!ht]
You are an expert Manim developer specializing in debugging and error resolution. Based on the provided implementation plan and Manim code, analyze the error message to provide a comprehensive fix and explanation.

\vspace{10pt}
\textbf{Implementation Plan:}
\{implementation\_plan\}

\vspace{10pt}
\textbf{Manim Code:}
\{manim\_code\}

\vspace{10pt}
\textbf{Error Message:}
\{error\_message\}

\vspace{10pt}
\textbf{Requirements:}
\begin{enumerate}
    \item Provide complete error analysis with specific line numbers where possible.
    \item Include exact instructions for every code change.
    \item Explain why the error occurred in plain language.
    \item ...
\end{enumerate}
\end{codeFixingBox}

\begin{evalBox}*[!ht]
You are a specialist in evaluating theorem explanation videos, known for giving clear and objective feedback. You will be given the transcript of a video. Your task is to evaluate and score the content of the video in several dimensions.

\vspace{10pt}
\textbf{Evaluation Criteria:}
\begin{enumerate}
    \item \textbf{Accuracy and Depth}
    \begin{itemize}
        \item Does the narration explain the theorem accurately?
        \item Does the video provide intuitive and/or rigorous explanations for why the theorem holds?
    \end{itemize}

    \item \textbf{Logical Flow}
    \begin{itemize}
        \item Does the video follow a clear and logical structure?
        \item Does the video present a coherent buildup of ideas?
    \end{itemize}
\end{enumerate}

\vspace{10pt}
\textbf{Scoring Instructions:} \\
Conduct a comprehensive evaluation and score each dimension from \text{0 to 1}: \\
(Score Descriptions)

\rule{\textwidth}{0.8pt} \\

You are tasked with analyzing and scoring a frame taken from a theorem explanation video. Note that you may not have the context of the video, so the captured frame may be a frame where some motion of visual elements is taking place. Your job is to assign a score from 1 to 5 for each criterion. Please provide a brief justification for your scores.

\vspace{10pt}
\textbf{Evaluation Criteria:}
\begin{enumerate}
    \item \textbf{Visual Relevance}
    \begin{itemize}
        \item Does the video frame align with the theorem's concepts and derivations?
    \end{itemize}
    
    \item \textbf{Element Layout}
    \begin{itemize}
        \item Are the visual elements well-placed and appropriately sized within the frame?
        \item Are the visual elements free of unintentional overlap?
        \item Is the visual information conveyed in the frame clear and easy to understand?
    \end{itemize}
\end{enumerate}
...

\rule{\textwidth}{0.8pt} \\

You are tasked with analyzing and scoring a chunk of a theorem explanation video. Note that you may not have the full context of the video. Your job is to assign a score from 1 to 5 for each criterion. Please provide a brief justification for your scores.

\vspace{10pt}
\textbf{Evaluation Criteria:}
\begin{enumerate}
    \item \textbf{Visual Consistency}
    \begin{itemize}
        \item Does the visual style remain consistent across frames?
        \item Are the motions and transitions smooth?
    \end{itemize}
\end{enumerate}
...
\end{evalBox}

\end{document}